\documentclass{article}
\usepackage{ijcai25}

\usepackage{times}
\usepackage{soul}
\usepackage{url}
\usepackage[hidelinks]{hyperref}
\usepackage[utf8]{inputenc}
\usepackage[small]{caption}
\usepackage{graphicx}
\usepackage{amsmath}
\usepackage{amsthm}
\usepackage{booktabs}
\usepackage{algorithm}
\usepackage{algorithmic}
\usepackage[switch]{lineno}

\usepackage{amssymb,amsfonts}
\usepackage{algorithmic}
\usepackage{tabularx}
\usepackage{multirow}
\usepackage{subfigure}

\newcommand{\figref}[1]{Figure~\ref{#1}}

\newcommand{\tabref}[1]{Table~\ref{#1}}

\newcommand{\model}{\text{LOVMM}}

\title{Language-Conditioned Open-Vocabulary Mobile Manipulation with Pretrained Models}

\author{
Shen Tan
\and
Dong Zhou$^{*}$\and
Xiangyu Shao\and
Junqiao Wang\and
Guanghui Sun\\
\affiliations
Harbin Institute of Technology\\
\emails
shentan@stu.hit.edu.cn,
dongzhou@hit.edu.cn,
xiangyushao@hit.edu.cn,
23S004019@stu.hit.edu.cn,
guanghuisun@hit.edu.cn
}

\begin{document}

\maketitle
\begin{abstract}
    Open-vocabulary mobile manipulation (OVMM) that involves the handling of novel and unseen objects across different workspaces remains a significant challenge for real-world robotic applications. In this paper, we propose a novel Language-conditioned Open-Vocabulary Mobile Manipulation framework, named LOVMM, incorporating the large language model (LLM) and vision-language model (VLM) to tackle various mobile manipulation tasks in household environments.\footnote{The source code, dataset, and supplementary material are available at: \url{https://github.com/shentan-shiina/LOVMM}.} Our approach is capable of solving various OVMM tasks with free-form natural language instructions (e.g. ``toss the food boxes on the office room desk to the trash bin in the corner'', and ``pack the bottles from the bed to the box in the guestroom''). Extensive experiments simulated in complex household environments show strong zero-shot generalization and multi-task learning abilities of LOVMM. Moreover, our approach can also generalize to multiple tabletop manipulation tasks and achieve better success rates compared to other state-of-the-art methods.\\
\end{abstract}
\section{Introduction}
    \begin{figure*}[ht]
        \centerline{\includegraphics[width=7 in]{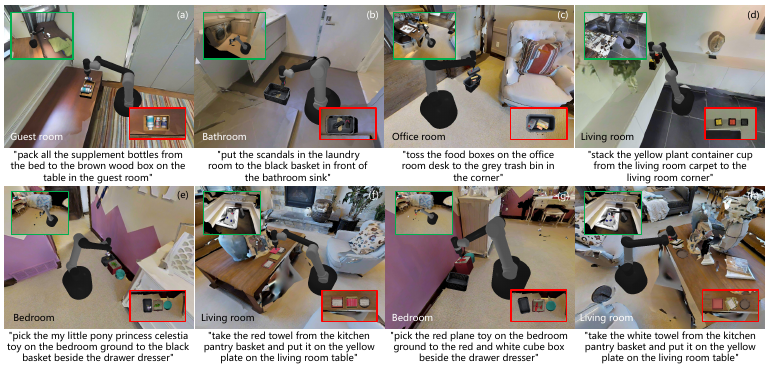}}
        \caption{Natural language-conditioned unseen OVMM tasks. We conduct large-scale experiments based on the CLIPort benchmark in simulated indoor household scenes for 16 OVMM tasks with over 35K steps of demonstrations (see Appendix~\ref{appendix:task_details_OVMM} for task details).}
        \label{fig:LOVMM_task_demo}
    \end{figure*}
    As one of the key capabilities for robotic home assistance, open-vocabulary mobile manipulation (OVMM), which leverages vision cameras to navigate in the environment and execute human-like actions to manipulate unseen objects, has attracted wide attention. It is crucial for addressing real-world challenges such as object sorting and rearrangement~\cite{zengRoboticPickplaceNovel2022c}, \cite{ganThreedworldTransportChallenge2022}, household cleanup~\cite{yanQuantifiableStratificationStrategy2021}, \cite{wuTidyBotPersonalizedRobot2023}, and human assistance~\cite{yenamandraHomeRobotOpenVocabularyMobile2024}, \cite{stoneOpenWorldObjectManipulation2023}.\\ 
    \indent Traditionally, robotic manipulation relies on vision-based methods that use explicit, object-centric representations, including poses, categories, and instance segmentations for perception~\cite{panTaxposeTaskspecificCrosspose2023a}, \cite{gengGapartnetCrosscategoryDomaingeneralizable2023a}, \cite{xieBestBothModes2020}. However, these approaches struggle with generalizing to unseen objects, as they often require specific training data for each scenario. Recently, end-to-end models that learn from expert demonstrations have emerged as promising alternatives~\cite{zengTransporterNetworksRearranging2021}, \cite{seitaLearningRearrangeDeformable2023}, \cite{gengRlaffordEndendAffordance2023a}. By leveraging visual observations without any explicit object information, these models are able to extract more generalizable representations across different tasks and zero-shot adapt to unseen scenarios. Yet, such methods are limited by the insufficient information provided by the single-modal data, or they may require goal images as instructions to adapt to new situations. In real-world scenarios, it is impractical to supply additional demonstrations or goal images for each new task. Thus, the model must possess the ability to open-vocabulary generalize to previously unseen tasks. An intuitive solution to this problem is grounding natural language in the manipulation policy. Natural language provides a direct interface for specifying targets and offers rich semantic information that is beneficial for more efficient learning. Although many efforts have been devoted to natural language conditioning for robotic manipulation~\cite{kamathMdetrmodulatedDetectionEndend2021a}, \cite{sharmaCorrectingRobotPlans2022c}, these models focus on explicit representations for seen objects, while natural language instructions are mainly used for target perception, rather than helping the model learn how to manipulate in an end-to-end manner.\\
    \indent Recent advancements in pretrained models, especially large language models (LLMs) and vision-language models (VLMs)~\cite{radfordLearningTransferableVisual2021a}, \cite{xueULIPLearningUnified2023}, have demonstrated zero-shot generalization capabilities across various robotic tasks. Notably, a number of works exploit the rich semantic information that lies in different modalities by combining natural language instructions with multi-view observations~\cite{goyalRVT2LearningPrecise2024}, 3D pointclouds~\cite{shridharPerceiveractorMultitaskTransformer2023a}, and action sequences \cite{brohanRT2VisionLanguageActionModels2023}. These models significantly improve generalization to novel objects, but they are often restricted to single workspaces or rely on simplified environments and predefined instructions, limiting their real-world applicability.\\
    \indent To address these challenges, we propose natural language-conditioned open-vocabulary mobile manipulation (\model), a framework that integrates the LLM for reasoning and VLMs for multimodal perception, enabling both open-vocabulary navigation and end-to-end manipulation with free-form natural language instructions. Specifically, we employ GPT-4~\cite{openaiGPT4TechnicalReport2024} to parse and reason for free-form natural language instructions and VLMaps~\cite{huangVisualLanguageMaps2023} to construct 3D vision-language maps for navigation. Following the architecture of CLIPort~\cite{shridharCliportWhatWhere2022}, we construct a two-stream model that fuses the semantic information from the vision-language representations of CLIP~\cite{radfordLearningTransferableVisual2021a} with the spatial information learned from the Transporter network~\cite{zengTransporterNetworksRearranging2021}. The fused features from both streams are exploited to predict 6-DoF manipulation poses, facilitating efficient 3D manipulation learning and generalization to complete unseen tasks across different workspaces.\\
    \indent We evaluate \model~in simulated household environments using a mobile suction gripper robot. Our experiments are built upon the CLIPort~\cite{shridharCliportWhatWhere2022} benchmark, including over 35K steps of demonstrations across 16 different seen and unseen language-conditioned tasks, each requires open-vocabulary navigation and cross-workspace manipulation ability, as shown in \figref{fig:LOVMM_task_demo}. \model~not only excels in multi-task learning for seen OVMM tasks, but also shows good zero-shot generalization performances for challenging unseen scenarios. In addition, the experiments further indicate that our model outperforms recent vision-based robotic manipulation methods in tabletop manipulation tasks, exhibiting more effective generalization capabilities. \\
    \indent The contributions of our work in this paper are summarized as follows:
    \begin{itemize}
    \item We propose a language-conditioned open-vocabulary mobile manipulation framework called \model, which enables the model to handle complex OVMM tasks with free-form natural language instructions in household environments.
    \item We present an end-to-end 6-DoF manipulation model that exploits the joint semantic and spatial information from multimodal input for learning accurate 3D manipulation efficiently.
    \item A variety of experiments based on the extended benchmark of OVMM tasks are conducted. The results show that \model~is able to zero-shot complete diverse OVMM tasks decently and achieves superior multi-task learning and generalization performances compared to recent vision-based manipulation models.
    \end{itemize}

\section{Related Work}
\begin{figure*}[!t]
    \centering
    \centerline{\includegraphics[width=6.5 in]{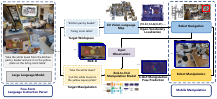}}
    \caption{Overview of LOVMM. We decompose OVMM as a series of open-vocabulary robot navigation and manipulation subtasks. Given a free-form natural language instruction, the LLM first parses the instruction to extract the target workspace and target manipulation description. The 3D vision-language map of the scene is then utilized to perform open-vocabulary localization for the robot to navigate to the specific workspace. Once the robot reaches the target location, the end-to-end manipulation model processes the target manipulation description along with RGB-D observations to predict a 6-DoF action pose for manipulation. By iteratively completing these subtasks, \model~enables the robot to perform complex OVMM tasks.}
    \label{fig:LOVMM_overview}
\end{figure*}
\subsection{Vision-based Robotic Manipulation}
    Perception for vision-based robotic manipulation has traditionally relied on object-centric representations such as pose estimation~\cite{panTaxposeTaskspecificCrosspose2023a}, keypoints~\cite{liuMOKAOpenVocabularyRobotic2024}, and dense descriptors~\cite{grafLearningDenseVisual2023a}. While these methods are effective, they typically require the manipulated objects to have rich texture details or complete 3D models to extract sufficient visual features. Thus, they struggle to generalize to unseen objects due to the lack of object-specific prior knowledge. \\
    \indent In contrast, recent advancements in deep learning-based methods demonstrate that leveraging visual observations directly, without prior object-centric information, helps the model to better understand perception-to-action concepts and learn more generalizable manipulation policies. The Transporter network~\cite{zengTransporterNetworksRearranging2021} finds the best object placement by cropping the image based on the sampled pick location and correlating the extracted deep visual features from both original and cropped RGB-D inputs to perform a template matching process. The correlation results naturally parameterize robot actions in a pick-and-place motion primitive. This formulation enables the network to learn manipulation skills efficiently but requires task-specific images to condition the policies, which limits its applicability in real-world scenarios. Further, Geng et al.~\cite{gengRlaffordEndendAffordance2023a} introduced a reinforcement learning framework that utilizes visual affordances to predict contact maps, providing a novel direction for end-to-end manipulation learning. Despite these advancements, the reliance solely on visual observations limits the perception capabilities of these methods, which makes them difficult to apply to real-world OVMM tasks. Our method, on the other hand, overcomes these limitations by integrating RGB-D images with natural language instructions as multi-modal input, which can better generalize to unseen scenarios and complete diverse OVMM tasks.
\subsection{Open-Vocabulary Mobile Manipulation}
    A number of prior works have explored how robots can solve various manipulation tasks, typically focusing on simple, single-workspace environments~\cite{zengTransporterNetworksRearranging2021}, \cite{seitaLearningRearrangeDeformable2023}, \cite{shridharCliportWhatWhere2022}. However, real-world robotic tasks often require complex mobile manipulation across multiple workspaces that involve both navigation and unseen object manipulation. For instance, a robot might need to retrieve an unfamiliar object from the kitchen and place it on the living room table, which is beyond the scope of traditional approaches. Such tasks are defined as OVMM and it remains an open problem \cite{yenamandraHomeRobotOpenVocabularyMobile2024}. Recent work~\cite{qiuOpenvocabularyMobileManipulation2024} proposed a two-stage framework based on 3D semantic mapping and pretrained models to decompose OVMM as a series of object fetching tasks, which achieves a decent success rate in a variety of real-world tasks. Moreover, the paper \cite{stoneOpenWorldObjectManipulation2023} focused on using various input modalities with VLM to solve open-world object manipulation and combined CoW~\cite{gadreCoWsPastureBaselines2022} to address open-vocabulary navigation and manipulation. Although these works provide innovative approaches for solving OVMM tasks, they struggle to generalize to different unseen environments and are restricted to simplified task setups that only work with seen objects or a single workspace. Our approach,~\model, advances this field by enabling zero-shot handling of diverse OVMM tasks that involve a wide range of complex, unseen environments with novel objects across different workspaces.
\subsection{Pretrained Models for Robotics}
    The advent of large pretrained models has sparked significant interest in applying their generalization capabilities to robotic tasks, including manipulation~\cite{zengTransporterNetworksRearranging2021}, navigation~\cite{gengRlaffordEndendAffordance2023a}, and even human assistance~\cite{kediaInteractTransformerModels2024}. Leveraging the strong reasoning and abstraction abilities of LLM, a number of researchers have introduced innovative approaches for grounding natural language and other different modalities into robotic learning. By decomposing high-level tasks into pretrained low-level skills with LLMs and using corresponding value functions to provide environment-specific knowledge, SayCan~\cite{ahnCanNotSay2022a} enables real-world long-horizon robotic task planning with natural language instructions.\\
    \indent In parallel, vision-language models (VLMs) that enable zero-shot capabilities by training on image-text pairs have also demonstrated impressive generalization performances in various tasks~\cite{liuGroundingDINOMarrying2024}, \cite{kirillovSegmentAnything2023}, \cite{yangDepthAnythingUnleashing2024}. By combining pretrained VLMs with imitation learning, some works have expanded traditional language-conditioned robotic manipulation paradigm~\cite{shridharPerceiveractorMultitaskTransformer2023a}, \cite{goyalRvtRoboticView2023a}, \cite{goyalRVT2LearningPrecise2024}. Previous work~\cite{shridharCliportWhatWhere2022} proposed a language-conditioned imitation learning agent that takes advantage of both CLIP~\cite{radfordLearningTransferableVisual2021a} and Transporter~\cite{zengTransporterNetworksRearranging2021} to learn general semantic concepts and precise spatial placement in few-shot settings. Nevertheless, it is constrained to 3-DoF manipulation tasks and it only focuses on manipulating in a fixed workspace with simple environment setups. Moreover, recent works that leverage vision-language-action models (VLAs) to directly learn generalizable robot actions are capable of tackling various manipulation tasks. However, these models often incorporate complex structures and require costly training~\cite{kim2024openvla}, \cite{team2024octo}. To this end, we propose a novel framework that exploits the strong language reasoning capabilities of GPT-4 for parsing free-form natural language instructions and the rich semantic information of pretrained VLMs for open-vocabulary navigation and manipulation. This enables efficient learning of 6-DoF manipulation skills and generalization of a wide range of OVMM tasks in complex, unseen environments.

\section{Proposed Method}
\begin{figure*}[!t]
    \centering
    \centerline{\includegraphics[width=7 in]{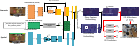}}
    \caption{Architecture of the proposed end-to-end manipulation model, which adopts a two-stream architecture to fuse the visual observations and natural language instruction of the pick-and-place manipulation tasks. The fused feature embeddings are then cropped and cross-correlated to produce a 2D affordance and are further exploited with multi-layer perceptrons to predict the 6-DoF manipulation pose.}
    \label{fig:LOVMM_architecture}
\end{figure*}
    In this section, we formulate the OVMM problem with free-form natural language instruction as input and describe \model~in detail. The overview of \model~is given in \figref{fig:LOVMM_overview}.
\subsection{Problem Formulation}
    The task of OVMM involves the mobile robot with vision cameras and the current environment. We formulate the problem as completing a series of subtasks as follows:
    \paragraph{Language Instruction Parsing.} Given a free-form natural language instruction ${\mathbf{L}}_{t}$ at each timestep $t$, it describes the target workspace ${\mathbf{l}}_{w_{t}}$ and target manipulation ${\mathbf{l}}_{m_{t}}$. We can parse the language instruction into such two parts ${\mathbf{L}}_{t} \to ({\mathbf{l}}_{w_{t}}, {\mathbf{l}}_{m_{t}})$, where each element is in text form. For example, an input instruction ``toss the food boxes on the office room desk'' can be parsed into the target workspace ``the office room desk'' and the target manipulation ``toss the food boxes''.
    \paragraph{Navigating to the Target Workspace.} The robot navigate to the position ${\mathbf{p}}_{t}$ of the current target workspace based on ${\mathbf{l}}_{w_{t}}$ and captures a visual observation ${\mathbf{o}}_{t}$ of the environment.
    \paragraph{Manipulating in the Target Workspace.} After reaching the target position, the problem can be considered as solving a tabletop pick-and-place manipulation subtask in the current workspace, which can be formulated as executing a manipulation policy $\pi$ that outputs robot actions ${\mathbf{a}}_{t}$: 
    \begin{equation}
    \pi({\mathbf{o}}_{t},{\mathbf{l}}_{m_{t}}) \to {\mathbf{a}}_{t}=({{\mathcal{T}}_{\text{pick}}},{{\mathcal{T}}_{\text{place}}}) \in \mathcal{A}
    \label{eq1}
\end{equation}
where ${{\mathcal{T}}_{\text{pick}}}$ and ${{\mathcal{T}}_{\text{place}}}$ are the poses of the robot end-effector for picking and placing actions, respectively. Both poses are defined in $\text{SE(3)}$ for 3D manipulation. The observation ${{\mathbf{o}}_{t}}$ is a top-down orthographic RGB-D projection of the workspace. The target manipulation description ${\mathbf{l}}_{m_{t}}$ specifies current manipulation task. When the current manipulation subtask is completed, the robot continues to navigate to the next given target workspace and manipulate again.\\
    \indent By repeating to finish the sequence of navigation and manipulation subtasks, the robot naturally solves the OVMM task.~In practice, we use input-action pairs ${{\zeta}_{i}}=\{({{\mathbf{o}}_{1}},{{\mathbf{l}}_{m_{1}}},{{\mathbf{a}}_{1}}),({{\mathbf{o}}_{2}},{{\mathbf{l}}_{m_{2}}},{{\mathbf{a}}_{2}}),\ldots\}$ to define each discrete-time tabletop manipulation instance, and the expert demonstration for each OVMM task can be presented as $\mathcal{D}_{i}=\{{({\mathbf{L}}_{1},{\mathbf{l}}_{w_{1}},{\mathbf{p}}_{1},{\zeta}_{1})},({{\mathbf{L}}_{2},{\mathbf{l}}_{w_{2}},{\mathbf{p}}_{2},{\zeta}_{2}}),\ldots\}$.
\subsection{Open-Vocabulary Navigation with Free-Form Natural Language Instruction}
    Given a free-form natural language instruction that specifies the OVMM task, \model~first uses GPT-4 as the free-form language instruction parser to interpret the target workspace ${\mathbf{l}}_{w_{t}}$ and target manipulation ${\mathbf{l}}_{m_{t}}$. Then, we follow the previous work~\cite{huangVisualLanguageMaps2023} and leverage the 3D reconstruction of current scene to construct a vision-language feature map matrix $Q \in \mathbb{R}^{\bar{H}\bar{W} \times C}$ using LSeg~\cite{liLanguagedrivenSemanticSegmentation2022} pixel embeddings, where $\bar{H}$ and $\bar{W}$ are the size of the predefined top-down grid map, $C$ is the length of the embedding vector. Each row of $Q$ represents the embedding of a pixel in the map. The parsed target workspace list ${\mathbf{l}}_{w_{t}}$ is encoded with the CLIP text encoder and organized into an embedding matrix $E \in \mathbb{R}^{M \times C}$, where $M$ represents the number of the target category. In this way, we can calculate the similarity between the given target workspace texts and the map pixels, and the highest ones indicate the most likely the pixels belong to the corresponding categories, which is formulated as:
    \begin{equation}
        {\mathbf{M}_{c}} = {\mathop{\text{argmax}}} Q \cdot E^{T}
    \end{equation}
where ${\mathbf{M}_{c}} \in \mathbb{R}^{\bar{H}\bar{W}}$, each element represents the label index of the target workspace categories. By choosing the most related pixels and reprojecting them to the original 3D scene map, we can localize each target workspace and obtain its position ${\mathbf{p}}_{t}$ for robot navigation.
\subsection{Natural Language-Conditioned End-to-End Manipulation}
\subsubsection{Learning 6-DoF Manipulation} 
After reaching the target workspace, the robot is ready to perform tabletop manipulation. We first construct our model with a similar template-matching approach based on the Transporter network~\cite{zengTransporterNetworksRearranging2021} to learn 2D planar manipulation, where  ${{\mathcal{T}}_{\text{pick}}},{{\mathcal{T}}_{\text{place}}} \in \text{SE(2)}$. Considering the above-mentioned pick-and-place policy $\pi$, we use fully convolutional networks (FCN) to model two action-value functions ${{\mathcal{Q}}_{\text{pick}}}$ and ${{\mathcal{Q}}_{\text{place}}}$. The first FCN ${{f}_{pick}}$ takes in ${{\gamma }_{t}}=( {\mathbf{o}}_{t},{\mathbf{l}}_{m_{t}})$ and outputs the pick action-value prediction ${{\mathcal{Q}}_{\text{pick}}} \in {{\mathbb{R}}^{H \times W}}$ that is used to calculate the pick action ${{\mathcal{T}}_{\text{pick}}}$:
\begin{equation}
    {{\mathcal{T}}_{\text{pick}}}=\underset{(u,v)}{\mathop{\text{argmax}}}\,{{\mathcal{Q}}_{\text{pick}}}( (u,v) \left| {{\gamma}_{t}} \right.)
    \label{eq2}
\end{equation}
where $(u,v)$ is the pixel location of the visual observation that can be mapped to the scene as a planar translation $(x,y)$ using camera-to-robot calibration. ${{\mathcal{T}}_{\text{pick}}} \sim (u,v) \in {{\mathbf{o}}_{t}}$ is the pick action at the corresponding location. The second and the third FCN $\psi$ and $\phi$ take in the same input ${{\gamma}_{t}}$ and outputs two feature embeddings of shape ${{\mathbb{R}}^{H \times W \times d}}$. Then, we crop the feature embedding from $\psi$ centered at ${{\mathcal{T}}_{\text{pick}}}$ as the query feature template to cross-correlate with the output key feature from $\phi$ to compute the place action-values ${{\mathcal{Q}}_{\text{place}}}$ and the corresponding place action ${{\mathcal{T}}_{\text{place}}}$:
\begin{equation}
    {{\mathcal{Q}}_{\text{place}}}(\Delta \tau \left| {{\gamma}_{t}} \right.,{{\mathcal{T}}_{\text{pick}}}) =((\psi ({{\gamma}_{t}})[{{\mathcal{T}}_{\text{pick}}}])*\phi ({{\gamma}_{t}}))[\Delta \tau]
    \label{eq3}
\end{equation}
\begin{equation}
    {{\mathcal{T}}_{\text{place}}}=\text{argma}{{\text{x}}_{\Delta \tau}}{{\mathcal{Q}}_{\text{place}}}(\left. \Delta \tau \right|{{\gamma}_{t}},{{\mathcal{T}}_{\text{pick}}})
    \label{eq4}
\end{equation}
where $\psi ({{\gamma}_{t}})[{{\mathcal{T}}_{\text{pick}}}]$ is the $c \times c$ partial crop of the feature embedding from $\psi$. Different from the original Transporter~\cite{zengTransporterNetworksRearranging2021}, we crop the feature embedding $\psi ({{\gamma}_{t}} )$ instead of cropping input observation ${{\mathbf{o}}_{t}}$ directly for a better receptive field. $\Delta \tau \in \text{SE(2)}$ represents the potential planar place pose, which is discretized into $k$ angles for the yaw rotation ${{\alpha}_{z}}$. In this work, we use $c=64$, $k=36$ and $d=3$. \\
    \indent With the current 2D actions ${{\mathcal{T}}_{\text{pick}}}$ and ${{\mathcal{T}}_{\text{place}}}$, we further leverage the rich spatial information embedded in the feature representations to learn 6-DoF manipulation. We apply a $1\times 1$ convolution after the output layers of $\psi$ and $\phi$ to adjust the feature channel dimension to ${d}'$. Then, by splitting the feature channels into three subsets, we utilize a separate cross-correlation and an MLP network $f(\cdot)$ for each subset to learn precise values for the remaining degrees-of-freedom, which can be formulated as follows:
\begin{equation}
    {\alpha}=f((({\psi}'({{\gamma}_{t}})[{{\mathcal{T}}_{\text{pick}}}])*{\phi}'({{\gamma}_{t}}))[{{\mathcal{T}}_{\text{place}}}])
    \label{eq5}
\end{equation}
where $\alpha$ represents each remaining degree-of-freedom: the roll angle ${{\alpha}_{x}}$, pitch angle ${{\alpha}_{y}}$, and height $z$. $\psi'$ and $\phi'$ are identical to $\psi$ and $\phi$ except for the additional $1 \times 1$ convolution, and we use ${d}'=24$. In this way, we can predict an accurate 6-DoF action pose for the manipulation target.
\begin{table*}[!htbp]
    \centering
    \begin{tabular}{
        >{\raggedright\arraybackslash}p{1.8cm}
        >{\centering\arraybackslash}p{0.65cm}
        >{\centering\arraybackslash}p{0.65cm}
        >{\centering\arraybackslash}p{0.65cm}
        >{\centering\arraybackslash}p{0.65cm}
        >{\centering\arraybackslash}p{0.65cm}
        >{\centering\arraybackslash}p{0.65cm}
        >{\centering\arraybackslash}p{0.65cm}
        >{\centering\arraybackslash}p{0.65cm}
        >{\centering\arraybackslash}p{0.65cm}
        >{\centering\arraybackslash}p{0.65cm}
        >{\centering\arraybackslash}p{0.65cm}
        >{\centering\arraybackslash}p{0.65cm}
        >{\centering\arraybackslash}p{0.65cm}
        >{\centering\arraybackslash}p{0.65cm}
        >{\centering\arraybackslash}p{0.65cm}
        }
    \toprule
    \multirow{3}{*}{Method} & \multicolumn{3}{c}{Task-A}  & \multicolumn{3}{c}{Task-B}  & \multicolumn{3}{c}{Task-C} & \multicolumn{3}{c}{Task-D}\\[1.5pt]
    \cmidrule(r){2-4} \cmidrule(r){5-7} \cmidrule(r){8-10} \cmidrule(r){11-13}
    & 1 & 10 & 100 & 1 & 10 & 100 & 1 & 10 & 100 & 1 & 10 & 100\\[1.5pt]
    \midrule
    LOVMM & 52.7 & 19.7 & \textbf{53.5} & \textbf{91.3} & 80.6 & 83.9 & 68.0 & \textbf{73.1} & 64.6 & 5.6 & 1.7 & \textbf{6.3}\\
    \midrule
    \multirow{3}{*}{Method}& \multicolumn{3}{c}{Task-E} &  \multicolumn{3}{c}{Task-F} & \multicolumn{3}{c}{Task-G} & \multicolumn{3}{c}{Task-H}\\[1.5pt]
    \cmidrule(r){2-4} \cmidrule(r){5-7} \cmidrule(r){8-10} \cmidrule(r){11-13}
    & 1 & 10 & 100 & 1 & 10 & 100 & 1 & 10 & 100 & 1 & 10 & 100\\[1.5pt]
    \midrule
    LOVMM & 63.0 & 34.0 & \textbf{69.0} & 43.1 & \textbf{82.9} & 82.5 & 59.0 & 34.0 & \textbf{62.0} & 44.3 & \textbf{72.7} & 63.9 \\                  
    \bottomrule
\end{tabular}
\caption{Seen OVMM tasks evaluation results.}
\label{tableI}
\end{table*}
\subsubsection{Two-stream Architecture} Specifically, all the FCNs ${{f}_{pick}}$, $\psi $ and $\phi $ are constructed based on a two-stream architecture~\cite{shridharCliportWhatWhere2022} to allow for natural language conditioning and joint semantic and spatial understanding, as shown in \figref{fig:LOVMM_architecture}. The spatial information stream uses an hourglass encoder-decoder model based on the Transporter ResNet~\cite{zengTransporterNetworksRearranging2021} network but with additional bottleneck layers for better spatial understanding, which takes in RGB-D visual observation ${{\mathbf{o}}_{t}}$ and outputs the feature embedding $\mathbf{d}_{t}^{(l)}$ at layer $l$. The semantic information stream leverages the pretrained CLIP ResNet-50~\cite{radfordLearningTransferableVisual2021a} image encoder to encode the RGB input ${{\mathbf{\tilde{o}}}_{t}}\to \mathbf{v}_{t}^{( 0 )}:{{\mathbb{R}}^{7\times 7\times 2048}}$, and uses skip-connected upsampling decoding layers to output feature tensors $\mathbf{v}_{t}^{( l-1 )}\to \mathbf{v}_{t}^{( l )}:{{\mathbb{R}}^{h\times w\times C}}$. To exploit the natural language instruction, the CLIP Transformer-based text encoder~\cite{radfordLearningTransferableVisual2021a} is utilized to get a language embedding ${\mathbf{l}}_{m_{t}}\to {{\mathbf{g}}_{t}}:{{\mathbb{R}}^{1024}}$. The language embedding is then downsampled and tiled with fully connected layers to produce ${{\mathbf{g}}_{t}}\to \mathbf{g}_{t}^{(l)}:{{\mathbb{R}}^{h\times w\times C}}$ such that the decoder feature embeddings of the semantic information stream can be conditioned through an element-wise product $\mathbf{v}_{t}^{(l)}\odot \mathbf{g}_{t}^{( l )}$. Then, the spatial and semantic information streams are fused with lateral connections that concatenate two feature tensors, and a $1\times 1$ convolution is applied to adjust the channel dimension of the output feature embedding, which is formulated as:
\begin{equation}
    [ \mathbf{v}_{t}^{(l)}\odot \mathbf{g}_{t}^{(l)};\mathbf{d}_{t}^{(l)} ]:{{\mathbb{R}}^{h\times w\times {{C}_{\mathbf{v}}}+{{C}_{\mathbf{d}}}}}\to {{\mathbb{R}}^{h\times w\times {{C}_{\mathbf{v}}}}}
    \label{eq6}
\end{equation}
where  ${{C}_{\mathbf{v}}}$ and ${{C}_{\mathbf{d}}}$ are the channel sizes of the semantic and spatial tensors.\\
\indent Finally, we train the end-to-end manipulation model through imitation learning from expert demonstrations $\mathcal{D}_{i}$. We first randomly sample an input-action pair ${{\zeta}_{i}}$ and then supervise the model with one-hot pixel encodings of the expert actions ${{Y}_{\text{pick}}}:{{\mathbb{R}}^{H\times W\times k}}$ and ${{Y}_{\text{place}}}:{{\mathbb{R}}^{H\times W\times k}}$ in an end-to-end manner. The model is trained with cross-entropy loss for 2D manipulation, which is defined as follows:
\begin{equation}
    \mathcal{L_{\text{2D}}}=-{{\mathbb{E}}_{{{Y}_{\text{pick}}}}}[ \text{log}{{\mathcal{V}}_{\text{pick}}} ]-{{\mathbb{E}}_{{{Y}_{\text{place}}}}}[\text{log}{{\mathcal{V}}_{\text{place}}}]
    \label{eq7}
\end{equation}
where ${{\mathcal{V}}_{\text{pick}}}=\text{softmax}( {{\mathcal{Q}}_{\text{pick}}}( \left. ( u,v ) \right|{{\gamma }_{t}} ) )$ and ${{\mathcal{V}}_{\text{place}}}=\text{softmax}( {{\mathcal{Q}}_{\text{place}}}( \left. \Delta \tau  \right|{{\gamma }_{t}},{{\mathcal{T}}_{\text{pick}}}) )$. For each remaining degree-of-freedom prediction for 3D manipulation, we use a Huber loss to train the MLP, which is formulated as follows:
\begin{equation}
    \mathcal{L_{\text{3D}}} = \begin{cases} 0.5 (\hat{\theta} - \theta)^2, & \text{if } |\hat{\theta} - \theta| < 1 \\ |\hat{\theta} - \theta| - 0.5, & \text{otherwise} \end{cases}
    \label{eq8}
\end{equation}
where $\hat{\theta}$ is the predicted value of each remaining degree-of-freedom, $\theta$ is the true value.

\section{Experiment}
    In this section, we evaluate the performance of \model~by conducting extensive simulated experiments in household environments. In addition, we further explore the effectiveness of our model by comparing it against baseline methods in natural language-conditioned open-vocabulary tabletop manipulation tasks. All models are trained on 4 NVIDIA RTX 4090 GPUs.
\subsection{\model~Performance for OVMM Tasks}
    We design 16 different natural language-conditioned OVMM tasks in various indoor scenes with household environments. The tasks are divided into 8 simpler seen tasks for training, and 8 more challenging unseen tasks for testing, as illustrated in \figref{fig:LOVMM_task_demo}. See Appendix~\ref{appendix:task_details_OVMM} for details on the task setup. The models are trained for 600K steps across all seen tasks using $n=1,10,100$ expert demonstrations in multi-task settings following CLIPort benchmark~\cite{shridharCliportWhatWhere2022}. Then we evaluate the models on 100 seen tasks and use the best validation model to test on 100 unseen tasks. The task success rate (TSR) is adopted to assess the model performance.
        \subsubsection{Performance for Seen OVMM Tasks} 
        The constructed OVMM seen tasks are shown in Appendix~\ref{appendix:task_details_OVMM} and the detailed evaluation results of all seen OVMM tasks are presented in \tabref{tableI}. It is obvious that LOVMM can decently solve most of the tasks, with the best $91.3\%$ TSR for \textit{Task-B}. Specifically, \model~trained with 100 task demonstrations outperforms other models in half of the tasks, with a performance of $53.5\%$ for \textit{Task-A}, which is over $30.0\%$ higher than the model trained with 10 demonstrations. We can also calculate the highest $60.7\%$ average TSR across all seen tasks of \model~using 100 demonstrations, showcasing its efficient multi-task learning and accurate manipulation abilities. Furthermore, when trained with limited demonstrations, \model~still shows decent performances, reaching $91.3\%$ and $59.0\%$ TSRs for \textit{Task-B} and \textit{Task-G}, respectively. LOVMM is able to reach a $53.4\%$ average TSR using only 1 seen task demonstration, which further demonstrates the efficient learning ability of \model~to grasp diverse manipulation skills. \\
        \indent However, a notable performance drop can be observed in some of the evaluated tasks when \model~is trained with 10 demonstrations. We hypothesize such a model behavior is caused by the imbalanced dataset and the random sampling strategy. See Appendix~\ref{appendix:limitations} for further discussion.
        \begin{table}[!tbp]
    \centering
    \begin{tabular}{
        >{\raggedright\arraybackslash}p{1.4cm}
        >{\centering\arraybackslash}p{0.3cm}
        >{\centering\arraybackslash}p{0.3cm}
        >{\centering\arraybackslash}p{0.3cm}
        >{\centering\arraybackslash}p{0.3cm}
        >{\centering\arraybackslash}p{0.3cm}
        >{\centering\arraybackslash}p{0.3cm}
        >{\centering\arraybackslash}p{0.3cm}
        >{\centering\arraybackslash}p{0.3cm}
        >{\centering\arraybackslash}p{0.3cm}
        >{\centering\arraybackslash}p{0.3cm}
        >{\centering\arraybackslash}p{0.3cm}
        >{\centering\arraybackslash}p{0.3cm}
        >{\centering\arraybackslash}p{0.3cm}
        >{\centering\arraybackslash}p{0.3cm}
        >{\centering\arraybackslash}p{0.3cm}
        }
    \toprule
    \multirow{1}*{Method} & \multicolumn{3}{c}{\multirow{1}*{Task-I}}  & \multicolumn{3}{c}{\multirow{1}*{Task-J}}  & \multicolumn{3}{c}{\multirow{1}*{Task-K}} & \multicolumn{3}{c}{\multirow{1}*{Task-L}} \\[1.5pt]
    \midrule
    LOVMM & \multicolumn{3}{c}{7.3} & \multicolumn{3}{c}{21.2} & \multicolumn{3}{c}{21.0} & \multicolumn{3}{c}{3.9} \\
    \midrule
    \multirow{1}{*}{Method} & \multicolumn{3}{c}{\multirow{1}*{Task-M}}  & \multicolumn{3}{c}{\multirow{1}*{Task-N}}  & \multicolumn{3}{c}{\multirow{1}*{Task-O}} & \multicolumn{3}{c}{\multirow{1}*{Task-P}} \\[1.5pt]
    \midrule
    LOVMM & \multicolumn{3}{c}{8.9} & \multicolumn{3}{c}{7.8} & \multicolumn{3}{c}{9.1} & \multicolumn{3}{c}{3.2} \\
    \bottomrule
\end{tabular}
\caption{Unseen OVMM tasks evaluation results.}
\label{tableII}
\end{table}
        \subsubsection{Performance for Unseen OVMM Tasks}
         Based on the evaluation performances for seen OVMM tasks, we choose \model~trained with 100 expert demonstrations to test on unseen OVMM tasks. The evaluation results are presented in \tabref{tableII}. See Appendix~\ref{appendix:task_details_unseen} for detailed evaluation results. Compared with seen tasks, the performances of \model~decrease as the unseen tasks are inherently more difficult and involved with diverse and challenging environments. It shows that our models can still generalize to complete many of them, reaching over $20.0\%$ TSR for tasks that require strong open-vocabulary manipulation capabilities, such as \textit{Task-J} and \textit{Task-K}. Notably, even though \textit{Task-I} requires generalization of unseen object categories and precise manipulation to solve, our model achieves the best performance of $7.3\%$ TSR, which shows the zero-shot open-vocabulary generalization and cross-workspace manipulation capabilities of \model. Our model also achieves nearly $10.0\%$ performance when generalizing to \textit{Task-O}, showcasing its 6-DoF manipulation learning ability. Even for extremely challenging tasks like \textit{Task-M}, which not only needs precise language parsing for open-vocabulary localization but also requires correct understanding of the corresponding pick-and-place targets, \model~manages to complete some instances. Covering all tasks, LOVMM is able to reach a $10.2\%$ average TSR using 100 task demonstration, which demonstrates that \model~has strong multi-task learning capabilities to leverage limited semantic information and manipulation concepts across different tasks efficiently to zero-shot generalize to unseen environments and novel object attributes.\\
        \indent In general, \model~is capable of learning manipulation skills efficiently and zero-shot generalizing to complete mobile manipulation in diverse, complex environments, providing a feasible solution for addressing a wide range of language-conditioned OVMM tasks with a one-for-all multi-task model.
        \begin{figure}[t]
            \centerline{\includegraphics[width=2.3 in]{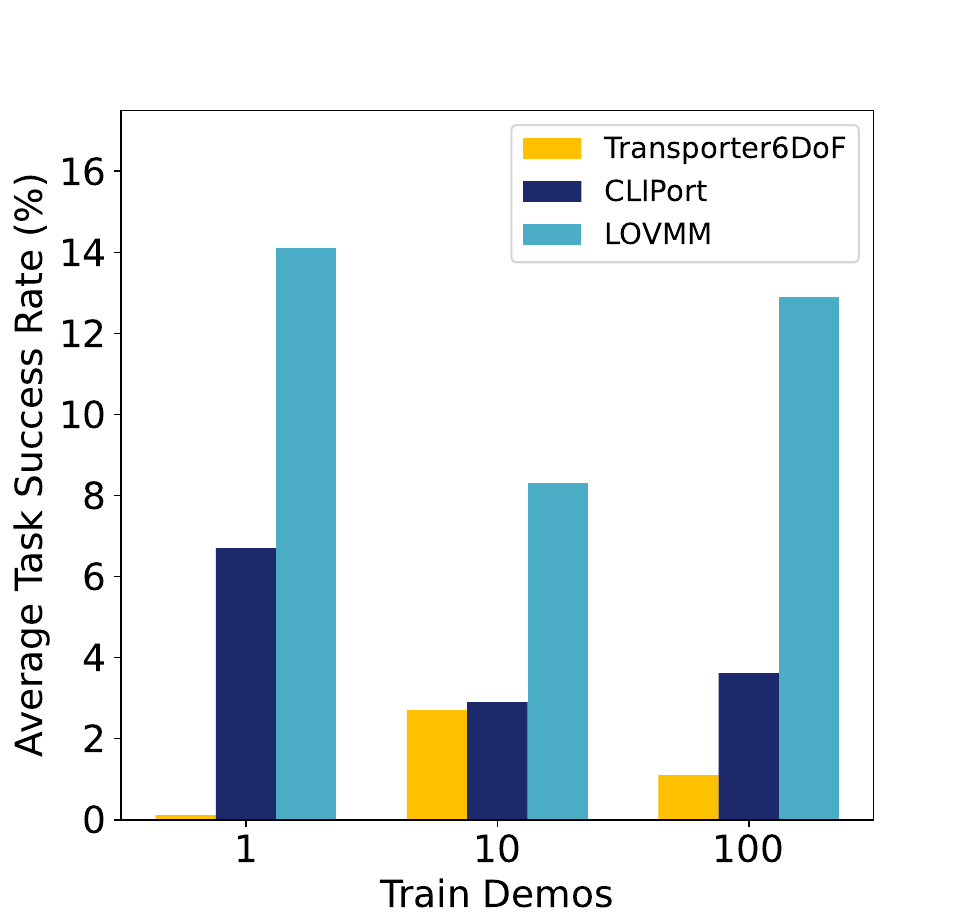}}
            \caption{Average task success rates for tabletop manipulation tasks.}
            \label{fig:LOVMM_OVMM_average}
        \end{figure}
\subsection{Performance Comparison for Tabletop Manipulation Tasks}
    To further validate the manipulation performance of our model, we compare our model against the image-conditioned Transporter with 6-DoF placing~\cite{zengTransporterNetworksRearranging2021} and language-conditioned CLIPort~\cite{shridharCliportWhatWhere2022} on 100 unseen tabletop manipulation tasks under the same multi-task training settings. The summarized average task success rates are shown in \figref{fig:LOVMM_OVMM_average}. See Appendix~\ref{appendix:task_details} for detailed evaluation results.\\
    \indent It is obvious in \figref{fig:LOVMM_OVMM_average} that our multi-task model performs exceptionally better than all the other baselines. To be specific, while Transporter6DoF can hardly reach only $0.1\%$ using 1 task demonstration, \model~reaches nearly $15.0\%$ TSR, surpassing the second-best CLIPort by more than double. As the number of training demonstrations increases, the average TSR of \model~first decreases to about $8.0\%$ but then recovers to over $12.0\%$ when trained with 100 demonstrations, which is 10 times better than the performance of Transporter6DoF. In contrast, the performances of CLIPort and Transporter6DoF show limited improvement as the number of expert demonstrations increases, with TSRs lower than $5.0\%$. These results further demonstrate that our LOVMM model has superior capabilities for efficient multi-task learning and adapting to novel unseen tasks in complex environments.
\subsection{Ablation Study}
    To evaluate the impact of various components of the proposed end-to-end manipulation model, we conduct a series of ablation studies. Specifically, we examine the effect of (i) removing data augmentation (details of data augmentation are presented in Appendix~\ref{appendix:data_augmentation}), (ii) cropping the input observation instead of the feature embedding, (iii) removing additional bottleneck layers, and (iv) using only 3-DoF manipulation. All ablation models are trained from scratch using 100 demonstrations and evaluated on 100 seen tasks. The results are summarized in \tabref{tableIII}. It clearly shows that each component contributes to the performance of \model. Removing data augmentation results in a substantial $19.1\%$ drop in average TSR, highlighting its crucial role in improving the model's learning efficiency and generalization. Moreover, cropping the observation directly may lead to the loss of receptive field, which makes the network less capable of perceiving complex environments, thus causing a performance decrease. Similarly, the additional bottleneck layers help the model extract richer perceptual information from the observations to grasp manipulation skills. The remaining degrees-of-freedom are also essential for the model to solve tasks that require 6-DoF placements such as \textit{Task-P}.\\
    \indent Overall, \model~demonstrates the ability to efficiently learn multi-task manipulation policies and zero-shot generalize to finish challenging OVMM tasks in unseen scenarios. The ablation results also highlight the importance of the proposed components in our end-to-end manipulation model, which are essential for LOVMM to achieve an excellent performance.
    \begin{table}[!tbp]
    \centering
    \begin{tabular}{
        >{\raggedright\arraybackslash}p{4.5cm}
        >{\centering\arraybackslash}p{0.7cm}
        >{\centering\arraybackslash}p{0.7cm}
        >{\centering\arraybackslash}p{0.7cm}
        >{\centering\arraybackslash}p{0.7cm}
        >{\centering\arraybackslash}p{0.7cm}
        >{\centering\arraybackslash}p{0.7cm}
        >{\centering\arraybackslash}p{0.7cm}
        }
    \toprule
    \multirow{1}{*}{Method} & \multicolumn{3}{c}{Average TSR}  \\[1.5pt]
    \midrule
    W/o data augmentation & \multicolumn{3}{c}{34.3} \\							
    Crop for input observation & \multicolumn{3}{c}{50.3} \\							
    No additional bottleneck layers & \multicolumn{3}{c}{48.6} \\							
    3-DoF manipulation & \multicolumn{3}{c}{47.9} \\													                                
    \midrule
    Original & \multicolumn{3}{c}{\textbf{53.4}} \\
    \bottomrule
\end{tabular}
\caption{Ablation Study on LOVMM for OVMM Tasks.}
\label{tableIII}
\end{table}

\section{Conclusion}
    In this paper, we formulate the OVMM task as completing a series of navigation and tabletop manipulation subtasks and propose a novel pretrained model-based natural language-conditioned open-vocabulary mobile manipulation framework, \model, incorporating an end-to-end manipulation model for efficient multi-task manipulation learning. Our framework enables free-form natural language instruction input and can learn generalizable policies to tackle diverse OVMM tasks that involve novel, unseen environments and objects. Extensive experiments simulated in household environments show the advancement of \model, and it achieves an overall better zero-shot generalization performance at solving tabletop manipulation tasks compared to other recent manipulation models.

\section*{Acknowledgements}
This study was supported by the National Natural Science Foundation of China through grant No. 62403162 and the InnoHK initiative of the Innovation and Technology Commission of the Hong Kong Special Administrative Region Government via the Hong Kong Centre for Logistics Robotics.

\bibliographystyle{named}
\bibliography{ijcai25}

\newpage
\appendix

\twocolumn[{
\centering
 \vspace{20pt}
\section*{\Large \centering Supplementary Material for LOVMM}
 \vspace{30pt}
 }]

\section{Appendix}
\subsection{Natural Language-conditioned OVMM Task Details}
\label{appendix:task_details_OVMM}
\begin{figure*}[t!]
    \centering
    \centerline{\includegraphics[width=7 in]{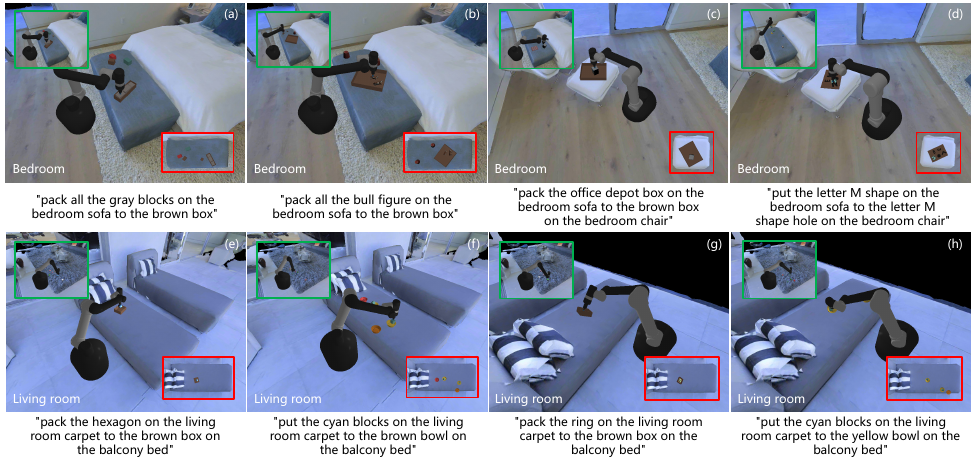}}
    \caption{Natural language-conditioned seen OVMM tasks.}
    \label{fig:LOVMM_OVMM_Tasks_Demos_Train}
\end{figure*}
\begin{figure*}[!t]
    \centering
    \centerline{\includegraphics[width=7 in]{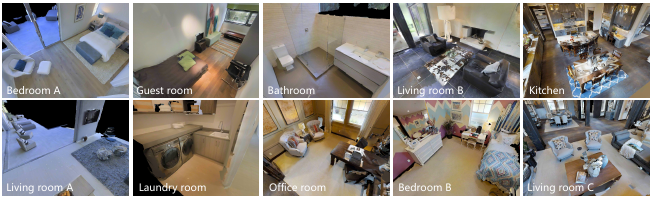}}
    \caption{Scenes for OVMM tasks.}
    \label{fig:LOVMM_Scenes}
\end{figure*}
\begin{figure*}[!ht]
    \centering
    \centerline{\includegraphics[width=5.4 in]{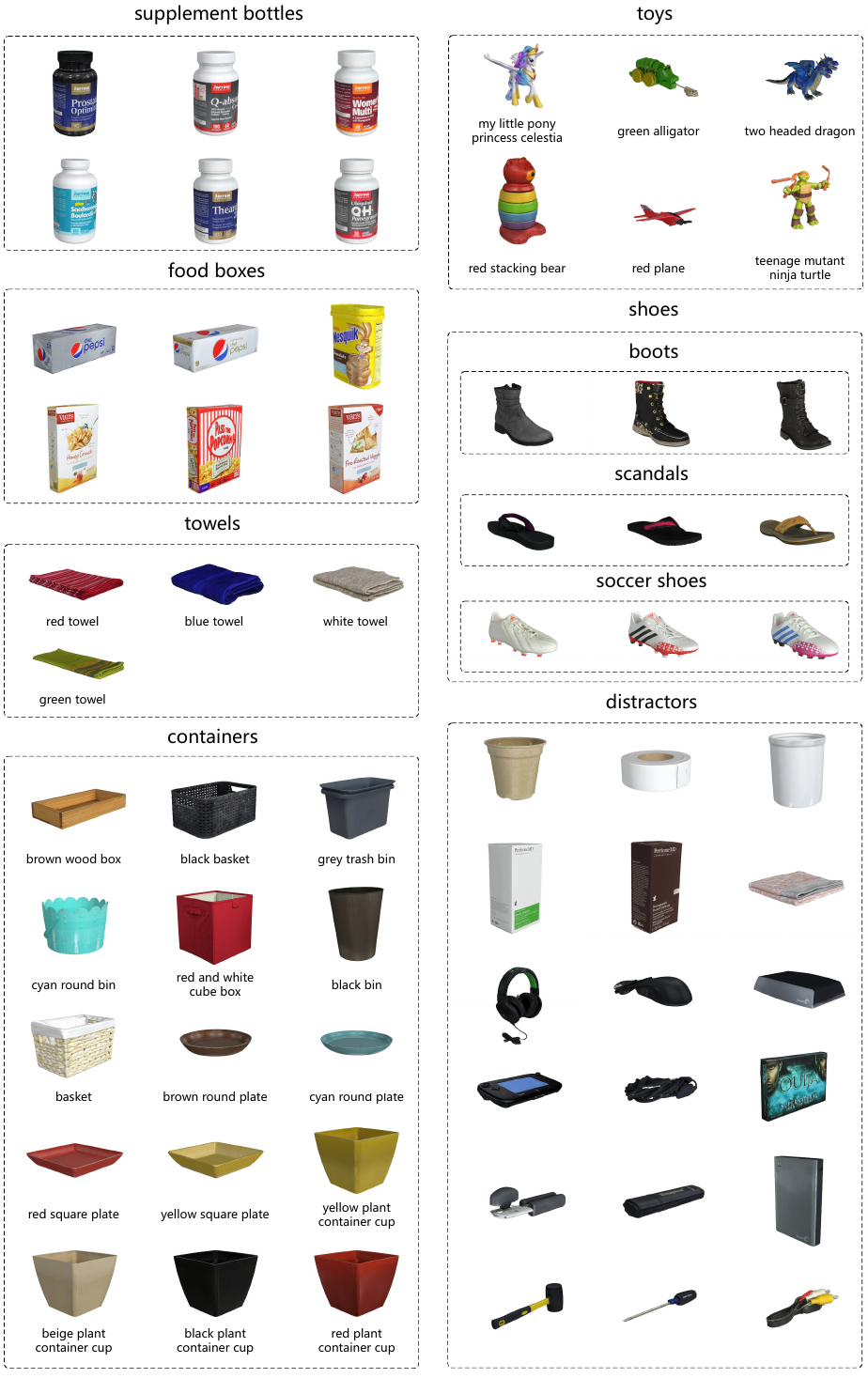}}
    \caption{Objects for OVMM tasks.}
    \label{fig:LOVMM_task_objects}
\end{figure*}
\begin{table*}[!ht]
    \centering
    \begin{tabular}{
        >{\raggedright\arraybackslash}p{1.2cm}
        >{\raggedright\arraybackslash}p{14.0cm}
        }
    \toprule
    \multirow{1}*{Task} & Natural Language Instruction\\[1.5pt]
    \midrule
	\multirow{1}*{Task-A} & ``pack all the \{color\} blocks on the bedroom sofa to the brown box''\\
    \midrule
	\multirow{1}*{Task-B} & ``pack all the \{google scan objects\} on the bedroom sofa to the brown box''\\
    \midrule
    \multirow{1}*{Task-C} & ``pack the \{google scan objects\} on the bedroom sofa to the brown box on the bedroom chair''\\
    \midrule
    \multirow{1}*{Task-D} & ``put the \{shapes\} on the bedroom sofa to the \{shapes\} hole on the bedroom chair''\\
    \midrule
	\multirow{1}*{Task-E} & ``put the \{shapes\} on the living room carpet to the brown box on the balcony bed''\\
    \midrule
    \multirow{1}*{Task-F} & ``put the \{color\} blocks on the living room carpet to the \{color\} bowl on the balcony bed''\\   
    \midrule
    \multirow{1}*{Task-G} & ``put the \{shapes\} on the living room carpet to the brown box on the balcony bed''\\
    \midrule
	\multirow{1}*{Task-H} & ``put the \{color\} blocks on the living room carpet to the \{color\} bowl on the balcony bed''\\
    \midrule
    \multirow{2}*{Task-I} & ``pack/put/pick all the supplement bottles from the bed to the brown wood box on the table in the guest room''\\
    \midrule
    \multirow{1}*{Task-J} & ``pack/put/pick the \{shoes\} in the laundry room to the black basket in front of the bathroom sink''\\
    \midrule
	\multirow{1}*{Task-K} & ``toss/put/pick the food boxes on the office room desk to the grey trash bin in the corner''\\
    \midrule
    \multirow{1}*{Task-L} & ``stack the \{containers\} from the living room carpet to the living room corner''\\
    \midrule
    \multirow{1}*{Task-M} & ``pick/pack/take the \{toys\} on the bedroom ground to the \{containers\} beside the drawer dresser''\\
    \midrule
	\multirow{2}*{Task-N} & ``take/put/pick the \{towels\} from the kitchen pantry basket and put it on the \{containers\} on the living room table''\\
    \midrule
    \multirow{1}*{Task-O} & ``pick/pack/take the \{toys\} on the bedroom ground to the \{containers\} beside the drawer dresser''\\
    \midrule
    \multirow{2}*{Task-P} & ``take/put/pick the \{towels\} from the kitchen pantry basket and put it on the \{containers\} on the living room table''\\
    \bottomrule
\end{tabular}
\caption{Natural language instructions for OVMM tasks.}
\label{tableV}
\end{table*}
We construct our natural language-conditioned OVMM tasks by extending the CLIPort benchmark~\cite{shridharCliportWhatWhere2022} into 10 different indoor scenes from the Matterport3D dataset~\cite{changMatterport3DLearningRGBData2017} using the Habitat simulator~\cite{szotHabitat20Training2021}, as shown in \figref{fig:LOVMM_Scenes}. To evaluate the multi-task learning and zero-shot generalizing capabilities of our model,  we build the seen tasks based on 2 simple scenes that only involve 4 workspaces with clean backgrounds such as the top of a flat sofa, while using 8 completely different scenes with 12 complex workspaces such as an uneven tabletop with messy background to construct the unseen tasks. Furthermore, the seen tasks are built based on the original benchmark without additional new objects, while the unseen OVMM tasks involve completely different novel objects from the Google Scanned Objects dataset~\cite{downsGoogleScannedObjects2022a}, as presented in \figref{fig:LOVMM_task_objects}. The unseen tasks also use more ambiguous, close-to-life language instructions with more difficult manipulation requirements. The natural language instructions used for each task are presented in \tabref{tableV}. For each task demonstration, the mobile robot is initialized in a random position in the scene. We name all the tasks with the corresponding scenes, target workspaces, and target manipulation descriptions. Unlike the original fixed workspace settings, we add an additional pick-and-place action step after the robot places the object from the previous workspace for more robust manipulation. For more details regarding the seen tasks, we refer the reader to the original CLIPort paper~\cite{shridharCliportWhatWhere2022}.
    \subsubsection{Bedroom-Sofa-Pack-Boxes (Task-A)}
    In this task, multiple blocks of different colors and sizes and a brown container box are placed on the sofa in the bedroom A scene. The robot needs to pack all the blocks of specified colors into the box to fill it tightly, as shown in \figref{fig:LOVMM_OVMM_Tasks_Demos_Train}(a). This task requires precise spatial and semantic understanding to handle different block sizes and colors. The task success rate (TSR) is defined as the volume of the specified blocks that are placed in the brown box divided by the total volume of all specified blocks in the scene.
    \subsubsection{Bedroom-Sofa-Pack-Google-Group (Task-B)} Multiple objects from the Google Scanned Objects dataset~\cite{downsGoogleScannedObjects2022a} and a brown container box are placed on the sofa in the bedroom A scene. The robot needs to pack all the specified objects into the box. This task requires strong open-vocabulary understanding capabilities to deal with different objects, as shown in \figref{fig:LOVMM_OVMM_Tasks_Demos_Train}(b). The TSR is defined as the volume of the specified objects that are placed in the container box divided by the total volume of all specified objects in the scene.
    \subsubsection{Bedroom-Sofa-Chair-Pack-Google-Seq (Task-C)} Similar to \textit{bedroom-sofa-pack-google-group}, multiple objects from the Google Scanned Objects dataset~\cite{downsGoogleScannedObjects2022a} are placed on the sofa, while a brown container box is placed on the chair in the bedroom A scene. The robot needs to pick up the specified objects from the sofa and place them into the container box on the chair. This task requires the model to perceive varying workspaces and generalize to handle different objects, as shown in \figref{fig:LOVMM_OVMM_Tasks_Demos_Train}(c). The TSR is defined as the volume of the specified objects that are placed in the container box divided by the total volume of all the specified objects in the scene.
    \subsubsection{Bedroom-Sofa-Chair-Assemble-Kits (Task-D)} Multiple shaped objects are placed on the sofa, while a brown board with different shaped holes is placed on the chair in the bedroom A scene. The robot needs to pick up objects of specified shapes and colors from the sofa and place them into the corresponding holes on the board. This task is particularly difficult as it requires not only precise spatial manipulation but also accurate semantic understanding across different workspaces to match different shapes and colors, as shown in \figref{fig:LOVMM_OVMM_Tasks_Demos_Train}(d). The TSR is defined as the number of correctly placed shaped objects divided by the total number of the shaped objects in the scene.
    \subsubsection{Livingroom-Carpet-Bed-Pack-Shapes (Task-E)} In this task, multiple shaped objects of randomized colors are placed on the carpet, and a brown container box is placed on the balcony bed in the living room A scene. The robot needs to pick up objects of specified shapes from the carpet and place them into the container box. This task requires strong environment perception capabilities to distinguish the shaped object from the complex carpet background, as shown in \figref{fig:LOVMM_OVMM_Tasks_Demos_Train}(e). The TSR is defined as the number of correctly placed shaped objects divided by the total number of the shaped objects in the scene.
    \subsubsection{Livingroom-Carpet-Bed-Put-Block-In-Bowl (Task-F)} Multiple blocks of different colors are placed on the carpet, and multiple bowls of different colors are placed on the balcony bed in the living room A scene. The robot needs to pick up blocks of specified colors from the carpet and place them into the bowls of the corresponding colors. This task requires accurate spatial manipulation to pick the block from the complex carpet background and demands the semantic understanding ability to match blocks and bowls of corresponding colors, as shown in \figref{fig:LOVMM_OVMM_Tasks_Demos_Train}(f). The TSR is defined as the number of correctly placed shaped objects divided by the total number of the shaped objects in the scene.
    \subsubsection{Livingroom-Carpet-Bed-Pack-Shapes-6dof (Task-G)} This task is identical to \textit{livingroom-carpet-bed-pack-shapes}, except that the target box is initialized with a random fixed 6-DoF pose within the angular range of $\pm \frac{\pi}{6}$ and height limit of $\text{10cm}$, which requires $\text{SE(3)}$ manipulation. The task example is presented in \figref{fig:LOVMM_OVMM_Tasks_Demos_Train}(g).
    \subsubsection{Livingroom-Carpet-Bed-Put-Block-In-Bowl-6dof (Task-H)} This task is identical to \textit{livingroom-carpet-bed-put-block-in-bowl}, except that each target bowl is initialized with a random fixed 6-DoF pose using the same angular range and height limit as \textit{livingroom-carpet-bed-pack-shapes-6dof}, which requires $\text{SE(3)}$ manipulation. The task example is presented in \figref{fig:LOVMM_OVMM_Tasks_Demos_Train}(h).
    \subsubsection{Guestroom-Bed-Table-Organize-Bottles (Task-I)} In this task, multiple different supplement bottles and some distractor objects are placed on the bed, while a brown wood container box is placed on the table in the guest room scene. The robot needs to pack all the supplement bottles into the box to fill it tightly, as shown in \figref{fig:LOVMM_task_demo}(a). We use ``supplement bottles'' to refer to all the targets without specifying each one in the language instruction. This task requires precise spatial and semantic understanding to handle different kinds of unseen bottles of different sizes. The TSR is defined as the volume of the supplement bottles that are placed in the brown wood box divided by the total volume of all the supplement bottles in the scene. 
    \subsubsection{Laundry-Bathroom-Basket-Pack-Shoes (Task-J)} Multiple kinds of different shoes are placed on the ground in the laundry room scene and a black basket is placed in front of the sink in the bathroom scene. The robot needs to pack all the specified shoes into the basket, as shown in \figref{fig:LOVMM_task_demo}(b). We use the category name such as ``boots'' to refer to all the target shoes without specifying each one in the language instruction. This task requires strong open-vocabulary generalization capabilities to distinguish different unseen kinds of shoes. The TSR is defined as the volume of the specified kind of shoes that are placed in the basket divided by the total volume of all specified shoes in the scene.
    \subsubsection{Officeroom-Table-Corner-Tidy-Food (Task-K)} In the office room scene, multiple different food boxes and some distractor objects are placed on the desk, while a grey trash bin is placed in the corner. The robot needs to put all the food boxes in the trash bin. We use ``food boxes'' to refer to all the targets and use ``toss'' instead of the seen ``pack'' in the language instruction to simulate real-life scenarios, as shown in \figref{fig:LOVMM_task_demo}(c). This task requires the model to correctly parse the language instruction and perceive the complex environment for accurate open-vocabulary manipulation. The TSR is defined as the volume of the food boxes that are placed in the trash bin divided by the total volume of all the food boxes in the scene.
    \begin{table*}[!ht]
    \centering
    \begin{tabular}{
        >{\raggedright\arraybackslash}p{2.2cm}
        >{\centering\arraybackslash}p{1.0cm}
        >{\centering\arraybackslash}p{1.0cm}
        >{\centering\arraybackslash}p{1.0cm}
        >{\centering\arraybackslash}p{1.0cm}
        >{\centering\arraybackslash}p{1.0cm}
        >{\centering\arraybackslash}p{1.0cm}
        >{\centering\arraybackslash}p{1.0cm}
        >{\centering\arraybackslash}p{1.0cm}
        >{\centering\arraybackslash}p{1.0cm}
        >{\centering\arraybackslash}p{1.0cm}
        >{\centering\arraybackslash}p{1.0cm}
        >{\centering\arraybackslash}p{1.0cm}
        >{\centering\arraybackslash}p{1.0cm}
        >{\centering\arraybackslash}p{1.0cm}
        >{\centering\arraybackslash}p{1.0cm}
        }
    \toprule[0.75pt]
    \multirow{2.5}*{Method} & \multicolumn{3}{c}{bed-table-organize-bottles}  & \multicolumn{3}{c}{basket-pack-shoes}  & \multicolumn{3}{c}{desk-corner-tidy-food}\\
    \cmidrule(r){2-4} \cmidrule(r){5-7} \cmidrule(r){8-10} 
    & 1 & 10 & 100 & 1 & 10 & 100 & 1 & 10 & 100\\[1.5pt]
    \toprule[0.75pt]
    Transporter6DoF & 0.0 & 0.0 & 0.1 & 0.0 & 0.0 & 8.0 & 1.0 & \textbf{21.5} & 0.7 \\
    CLIPort & 24.5 & 7.1 & 0.0 & \textbf{21.5} & 11.9 & 0.0 & 0.0 & 0.7 & 3.0 \\
    LOVMM & \textbf{55.9} & \textbf{26.7} & \textbf{8.3} & 19.7 & \textbf{21.3} & \textbf{23.2} & \textbf{7.9} & 8.3 & \textbf{26.6} \\
	
    \toprule[0.75pt] 
    \multirow{2.5}*{Method} & \multicolumn{3}{c}{carpet-corner-stack-cups} & \multicolumn{3}{c}{ground-drawer-sort-toys}  & \multicolumn{3}{c}{basket-table-put-towels} \\
    \cmidrule(r){2-4} \cmidrule(r){5-7} \cmidrule(r){8-10} 
    & 1 & 10 & 100 & 1 & 10 & 100 & 1 & 10 & 100  \\[1.5pt]
    \toprule[0.75pt]
    Transporter6DoF & 0.0 & 0.0 & 0.0 & 0.0 & 0.0 & 0.2 & 0.0 & 0.0 & 0.0 \\
    CLIPort & 1.3 & 0.1 & 3.0 & 3.2 & 1.1 & 7.9 & 2.7 & 1.8 & \textbf{10.5} \\
    LOVMM & \textbf{7.3} & \textbf{1.3} & \textbf{8.3} & \textbf{5.3} & \textbf{2.7} & \textbf{11.2} & \textbf{6.2} & \textbf{3.1} & 9.6 \\
    
    \toprule[0.75pt] 
    \multirow{2.5}*{Method} & \multicolumn{3}{c}{ground-drawer-sort-toys-6dof} &  \multicolumn{3}{c}{basket-table-put-towels-6dof} & \multicolumn{3}{c}{Average TSR}\\
    \cmidrule(r){2-4} \cmidrule(r){5-7} \cmidrule(r){8-10}
    & 1 & 10 & 100 & 1 & 10 & 100 & 1 & 10 & 100 \\[1.5pt]
    \toprule[0.75pt]
    Transporter6DoF & 0.0 & 0.0 & 0.0 & 0.0 & 0.0 & 0.0 & 0.1 & 2.7 & 1.1 \\
    CLIPort & 0.3 & 0.4 & 3.2 & 0.0 & 0.0 & 1.3 & 6.7 & 2.9 & 3.6 \\
    LOVMM & \textbf{6.3} & \textbf{1.2} & \textbf{10.2} & \textbf{3.9} & \textbf{2.0} & \textbf{5.6} & \textbf{14.1} & \textbf{8.3} & \textbf{12.9} \\                  
    \toprule[0.75pt]
\end{tabular}
\caption{Tabletop manipulation tasks evaluation results.}
\label{tableIV}
\end{table*}
    \subsubsection{Livingroom-Carpet-Corner-Stack-Cups (Task-L)} Multiple plant container cups of different colors are placed on the carpet and the corner in the living room B scene. The robot needs to pick up the specified cups from the carpet and stack them on top of the cups of corresponding colors in the corner. This task is particularly difficult as it requires not only precise spatial manipulation but also accurate perceiving capabilities to pick from the complex carpet background and match different cups across different workspaces, as shown in \figref{fig:LOVMM_task_demo}(d). The TSR is defined as the number of correctly placed cups divided by the total number of the specified cups in the scene.
    \subsubsection{Bedroom-Ground-Drawer-Sort-Toys (Task-M)} In this task, multiple different toys and some distractor objects are placed on the ground, while three different containers are placed beside the drawer dresser in the bedroom B scene. The robot needs to pick up the specified toys and place them into the specified container. This task requires strong open-vocabulary generalization and semantic understanding capabilities to distinguish and match the specified unseen toys and containers, as shown in \figref{fig:LOVMM_task_demo}(e). The TSR is defined as the volume of the specified toys that are correctly placed in the corresponding containers divided by the total volume of all the toys in the scene.
    \subsubsection{Kitchen-Livingroom-Basket-Table-Put-Towels (Task-N)} Multiple towels of different sizes and colors are placed in the pantry basket in the kitchen scene, and multiple plates of different sizes and colors are placed on the desk table in the living room C scene. The robot needs to pick up towels of specified colors from the pantry basket and place them into the plates of the specified colors, as shown in \figref{fig:LOVMM_task_demo}(f). This task not only requires accurate spatial manipulation but also demands the semantic generalization ability to match unseen towels and plates. Moreover, it also requires accurate language parsing for the robot to open-vocabulary navigate to the correct workspaces. The TSR is defined as the volume of the specified towels that are correctly placed divided by the total volume of all the specified towels in the scene.
    \subsubsection{Bedroom-Ground-Drawer-Sort-Toys-6dof (Task-O)} This task is identical to \textit{bedroom-ground-drawer-sort-toys}, except that the target container is initialized with a random fixed 6-DoF pose within the angular range of $\pm \frac{\pi}{4}$ and height limit of $\text{10cm}$, which requires $\text{SE(3)}$ manipulation. The task example is presented in \figref{fig:LOVMM_task_demo}(g).
    \subsubsection{Kitchen-Livingroom-Basket-Table-Put-Towels-6dof (Task-P)} This task is identical to \textit{kitchen-livingroom-basket-table-put-towels}, except that each target bowl is initialized with a random fixed 6-DoF pose using the same angular range and height limit as \textit{bedroom-ground-drawer-sort-toys-6dof}, which requires $\text{SE(3)}$ manipulation. The task example is presented in \figref{fig:LOVMM_task_demo}(h).           
    
\begin{figure*}[!t]
    \centering
    \centerline{\includegraphics[width=5 in]{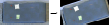}}
    \caption{Data augmentation for training observation samples.}
    \label{fig:LOVMM_data_aug}
\end{figure*}
\subsection{Evaluation Results for Unseen OVMM Tasks}
    To further validate the performance of LOVMM, we implement the tabletop manipulation baselines with manually annotated fixed navigation routes (FNR) to complete unseen OVMM tasks. All models are trained on the same set of 100 demonstrations from seen tasks. Detailed evaluation results are presented in \tabref{tableVI}. It shows that LOVMM outperforms both baselines across all tasks with a notable margin. Specifically, LOVMM achieves over $18.0\%$ higher TSR than FNR+CLIPort for \textit{Task-K} and generalizes nearly 3 times better than FNR+Transporter6DoF for \textit{Task-J}. Moreover, our model is capable of solving challenging tasks such as \textit{Task-L}, where both baselines fail to complete even a single instance. Such results further demonstrate the strong manipulation learning and generalizing abilities of LOVMM in solving complex OVMM tasks.
\label{appendix:task_details_unseen}
\begin{table}[!tbp]
    \centering
    \setlength\tabcolsep{2.5pt}
    \begin{tabular}{
        >{\raggedright\arraybackslash}p{3.4cm}
        >{\centering\arraybackslash}p{0.15cm}
        >{\centering\arraybackslash}p{0.15cm}
        >{\centering\arraybackslash}p{0.15cm}
        >{\centering\arraybackslash}p{0.15cm}
        }
    \toprule
    \multirow{1}*{Method} & \multicolumn{1}{c}{\multirow{1}*{Task-I}}  & \multicolumn{1}{c}{\multirow{1}*{Task-J}}  & \multicolumn{1}{c}{\multirow{1}*{Task-K}} & \multicolumn{1}{c}{\multirow{1}*{Task-L}} \\[1.5pt]
    \midrule
	FNR+Transporter6DoF & \multicolumn{1}{c}{0.0} & \multicolumn{1}{c}{7.6} & \multicolumn{1}{c}{0.6} & \multicolumn{1}{c}{0.0} \\
	FNR+CLIPort & \multicolumn{1}{c}{0.0} & \multicolumn{1}{c}{0.0} & \multicolumn{1}{c}{2.8} & \multicolumn{1}{c}{0.0} \\
    LOVMM & \multicolumn{1}{c}{\textbf{7.3}} & \multicolumn{1}{c}{\textbf{21.2}} & \multicolumn{1}{c}{\textbf{21.0}} & \multicolumn{1}{c}{\textbf{3.9}} \\
    \midrule
    \multirow{1}{*}{Method} & \multicolumn{1}{c}{\multirow{1}*{Task-M}}  & \multicolumn{1}{c}{\multirow{1}*{Task-N}}  & \multicolumn{1}{c}{\multirow{1}*{Task-O}} & \multicolumn{1}{c}{\multirow{1}*{Task-P}} \\[1.5pt]
    \midrule
	FNR+Transporter6DoF & \multicolumn{1}{c}{0.0} & \multicolumn{1}{c}{0.0} & \multicolumn{1}{c}{0.0} & \multicolumn{1}{c}{0.0} \\
	FNR+CLIPort & \multicolumn{1}{c}{7.5} & \multicolumn{1}{c}{6.5} & \multicolumn{1}{c}{3.0} & \multicolumn{1}{c}{1.3} \\
    LOVMM & \multicolumn{1}{c}{\textbf{8.9}} & \multicolumn{1}{c}{\textbf{7.8}} & \multicolumn{1}{c}{\textbf{9.1}} & \multicolumn{1}{c}{\textbf{3.2}} \\
    \bottomrule
\end{tabular}
\caption{Unseen OVMM tasks evaluation results.}
\label{tableVI}
\end{table}
\subsection{Evaluation Results for Tabletop Manipulation Tasks}
    We evaluate the models on tabletop manipulation tasks by using the same target workspace observation for all models to complete unseen OVMM tasks. In this way, the tasks are simplified to completing tabletop manipulation at each workspace without any navigation requirements. The tasks are named the same as OVMM tasks but without the scene names. The detailed evaluation results are presented in \tabref{tableIV}. It shows that \model~outperforms all the other models in $21/24 = 87.5\%$ of the evaluated tasks. Specifically, LOVMM performs exceptionally well compared with the baselines in many tasks, achieving a best $55.9\%$ TSR for \textit{bed-table-organize-bottles} with only 1 demonstration, and a $26.6\%$ performance for \textit{desk-corner-tidy-food}. For tasks that are more challenging, CLIPort struggles to achieve less than $5.0\%$ TSR with limited expert demonstrations and Transporter6DoF is unable to solve most of the tasks. On the other hand, the $6.3\%$ performance of LOVMM in the challenging \textit{ground-drawer-sort-toys-6dof} that involves both complex environments and 6-DoF manipulation compared with the baselines' near-zero TSR also indicates that our model is capable of adapting to unseen workspace environments and open-vocabulary generalize to different object attributes.
\label{appendix:task_details}

\subsection{Data Augmentation}
    Following the data augmentation setting in the original CLIPort implementation~\cite{shridharCliportWhatWhere2022}, we apply random $\text{SE(2)}$ transformations to the training observation samples for better spatially-equivariant representation learning. Additionally, we apply multiple object lighting settings, adjust brightness, and add random noises and Gaussian blur to the observation samples to enhance the model's generalization ability, as shown in \figref{fig:LOVMM_data_aug}.
\label{appendix:data_augmentation}

\subsection{Limitations}
    \subsubsection{Imbalanced Dataset} Our proposed natural language-conditioned OVMM tasks involve a wide range of scenarios, covering single-step manipulation to long-horizon tasks. As a result, the dataset is heavily imbalanced. In order to have a fair and direct comparison with the baseline methods, we adopt the original random sampling strategy used by CLIPort, which inevitably introduces learning bias across tasks. In this case, more demonstrations may lead to sparser task coverage, resulting in degraded manipulation performance. Future work could incorporate weighted sampling methods such as \cite{team2024octo} to address this issue.
    \subsubsection{Simplified Task Settings} Although LOVMM's simple and low-cost settings offer a practical and scalable solution for real-world deployment, real-world tasks often require multi-view observations and continuous 6-DoF actions, such as pouring a cup of water into the container. However, multi-view inputs may surge computational cost, and setting up multiple cameras is non-trivial. Future work could explore more flexible camera setups and action settings. Furthermore, in real-world situations, navigation and manipulation are often related (e.g., going to the bedroom needs opening the door first), so developing an entangled pipeline for OVMM would be a promising direction.    
\label{appendix:limitations}

\end{document}